\documentclass[letterpaper]{article} 
\usepackage{aaai23}  
\usepackage{times}  
\usepackage{helvet}  
\usepackage{courier}  
\usepackage[hyphens]{url}  
\usepackage{graphicx} 
\usepackage{natbib}  
\usepackage{caption} 
\usepackage[ruled,vlined,linesnumbered]{algorithm2e}
\usepackage{xcolor}
\usepackage{subcaption}
\usepackage{newfloat}
\usepackage{listings}
\DeclareCaptionStyle{ruled}{labelfont=normalfont,labelsep=colon,strut=off} 
\usepackage[page]{appendix}

\usepackage{hyperref}

\title{
TransPath: Learning Heuristics For Grid-Based Pathfinding via Transformers
}
\author {
    Daniil Kirilenko, \textsuperscript{\rm 1} 
    Anton Andreychuk, \textsuperscript{\rm 2} 
    Aleksandr Panov, \textsuperscript{\rm 1, 2}
    Konstantin Yakovlev \textsuperscript{\rm 1, 2}
}
\affiliations {
    \textsuperscript{\rm 1} Federal Research Center for Computer Science and Control of Russian Academy of Sciences, Moscow, Russia\\
    \textsuperscript{\rm 2} AIRI, Moscow, Russia\\
    anedanman@gmail.com, andreychuk@airi.net, panov@airi.net, yakovlev@isa.ru
}

\begin{document}

\maketitle

\begin{abstract}
Heuristic search algorithms, e.g. A*, are the commonly used tools for pathfinding on grids, i.e. graphs of regular structure that are widely employed to represent environments in robotics, video games etc. Instance-independent heuristics for grid graphs, e.g. Manhattan distance, do not take the obstacles into account and, thus, the search led by such heuristics performs poorly in the obstacle-rich environments. To this end, we suggest learning the instance-dependent heuristic proxies that are supposed to notably increase the efficiency of the search. The first heuristic proxy we suggest to learn is the correction factor, i.e. the ratio between the instance independent cost-to-go estimate and the perfect one (computed offline at the training phase). Unlike learning the absolute values of the cost-to-go heuristic function, which was known before, when learning the correction factor the knowledge of the instance-independent heuristic is utilized. The second heuristic proxy is the path probability, which indicates how likely the grid cell is lying on the shortest path. This heuristic can be utilized in the Focal Search framework as the secondary heuristic, allowing us to preserve the guarantees on the bounded sub-optimality of the solution. We learn both suggested heuristics in a supervised fashion with the state-of-the-art neural networks containing attention blocks (transformers). We conduct a thorough empirical evaluation on a comprehensive dataset of planning tasks, showing that the suggested techniques \emph{i}) reduce the computational effort of the A* up to a factor of $4$x while producing the solutions, which costs exceed the costs of the optimal solutions by less than $0.3$\% on average; \emph{ii}) outperform the competitors, which include the conventional techniques from the heuristic search, i.e. weighted A*, as well as the state-of-the-art learnable planners.

The project web-page is: \url{https://airi-institute.github.io/TransPath/}.

\end{abstract}

\section{Introduction}

Path planning for a mobile agent in the static environment is a fundamental problem in AI that is often framed as a graph search problem. Within this approach, first, an agent's workspace is discretized to a graph, second, a search algorithm is invoked on this graph to find a path from start to goal. Perhaps $2^k$-connected grids~\cite{rivera20202} are the most widely used graphs for path planning in a variety of applications (robotics, video games, etc.).

\begin{figure}[t]
  \centering
  \includegraphics[width=1\linewidth]{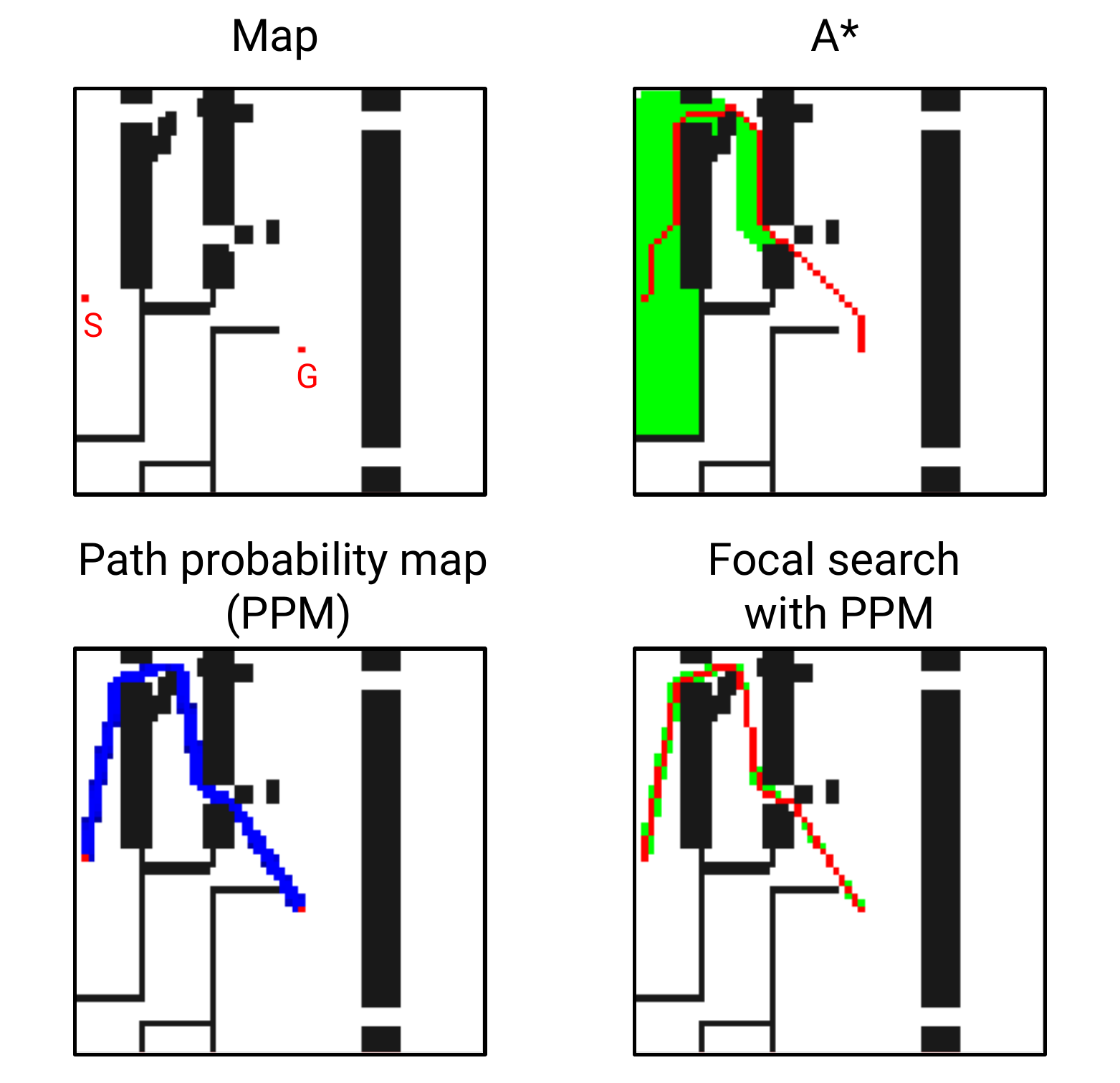}
\caption{The difference between A* and our approach. Expanded nodes are shown in green, while path in red. Blue regions are predicted by the neural network to contain path cells with high probability.}
\label{fig:visual-abstract}
\end{figure}

Path planning on a grid is commonly accomplished by a heuristic search algorithm, e.g. A*~\cite{hart1968formal} or one of its numerous modifications. Performance of such algorithms is heavily dependent on the input heuristic that comes in the form of a function that estimates the cost of the path to the goal for each node of the graph (\emph{cost-to-go heuristic}). If the heuristic is perfect, i.e. for every node its value equals the cost of the shortest path, a search algorithm, guided by it, explores only the nodes that lie on one of the minimum-cost paths. However, such a perfect heuristic is instance-dependent and cannot be encoded in the closed-loop form. In practice, instance-independent heuristics, e.g. Manhattan distance, are typically used for grid-based path planning. These heuristics do not take obstacles into account, and, consequently, perform poorly in the obstacle-rich environments.

One of the recent and promising approaches to automated construction of the instance-dependent heuristics (and for path planning in general) is utilizing machine learning, and specifically deep learning~\cite{Speck2020,Janner2022}. As grids can be viewed as the binary images, it is appealing to utilize the recent advances in convolutional neural networks (CNNs)~\cite{DBLP:journals/corr/abs-1905-11946, DBLP:journals/corr/abs-1710-05941} to extract the informative features from the image representations of the pathfinding problems and embed these features to the heuristic search algorithm. For example, in~\cite{takahashi2019learning} it was suggested to learn perfect cost-to-go heuristic in a supervised fashion.  
In a more recent study~\cite{yonetani2021path}, a more involved approach was introduced when a matrix-based A* was proposed and used for learning. Thus, the deep neural network model was trained end-to-end. In that work not the conventional cost-to-go heuristic was predicted but rather the additional cost that was assigned to each grid cell with the intuition that unpromising nodes will be assigned a high cost by the neural network. Consequently, at the planning phase, the search will avoid the cells with the high costs.

In this work, we follow the described paradigm and further examine the ways of how heuristic search can benefit from state-of-the-art deep learning techniques in the context of the grid-based path planning. The distinguishable features of our work are as follows. We consider 8-connected grids with non-uniform costs (i.e., diagonal moves cost more than the cardinal ones), unlike the previous works that considered unit cost domains. We suggest learning the novel heuristic proxies for the problem at hand. Instead of learning to predict the values of the perfect cost-to-go heuristic, we suggest learning the \emph{correction factor} of the heuristic function, which is the ratio between the instance-independent heuristic and the perfect heuristic. Thus a correction factor embed information about both of these heuristics. Our empirical evaluation confirms that learning the correction factor leads to a notably better performance than learning the absolute values of the  conventional cost-to-go heuristic.

We also suggest learning the \emph{path probability map}, which assigns to each grid cell the probability of belonging to the shortest path. This can be used as a secondary heuristic in the bounded sub-optimal search algorithm -- Focal Search~\cite{pearl1982studies}. Thus, we are able to preserve the theoretical guarantees on a sub-optimality bound of the constructed solution, while speeding up the search, as our experiments show. 

To learn the correction factor and path probabilities, we utilize supervised deep learning. We employ a neural network model, that is a combination of the convolutional encoder-decoder with the attention blocks~\cite{vaswani2017attention} (the so-called transformers). Such a combination allows the neural network to capture and ``reason about'' both the local features of a given map (corners of obstacles, passages etc.), and the relations between them, e.g. ``there is a passage between two regions of interest''.

To evaluate the suggested techniques, a comprehensive dataset of challenging planning tasks was created that extends the dataset previously used in closely related works~\cite{yonetani2021path}. We compare our approach with the competitors that include both the deep learning techniques and the traditional ones and demonstrate its superiority in terms of the computational effort and solution cost. Overall we were able to reduce the computational effort compared to A* up to a factor of $4$x while producing the solutions, which costs exceed the costs of the optimal solutions by less than $0.3$\% on average.

\section{Related Work}

The following two lines of research are especially relevant to our work: \emph{i}) techniques that trade-off optimality for computational efficiency; \emph{ii}) combining data-driven machine learning with the heuristic search for grid-based pathfinding.

\paragraph{Trading-off optimality for computational efficiency.} A classic technique for such trading-off, widely used in practice, is running A* with the heuristic function multiplied by a constant $w \geq 1$ -- the so-called weighted A* (WA*)~\cite{pohl1970heuristic}. It guarantees finding the solutions that have a cost that is at most $w$ times the optimal solution cost. Thus, the solution is \emph{bounded sub-optimal}. When time permits, a series of searches can be performed, each one with the decreased value of $w$ -- anytime search~\cite{likhachev2004ara,hansen2007anytime}. Another well-known technique for bounded sup-optimal search is Focal Search~\cite{pearl1982studies}, whose anytime versions are also known~\cite{cohen2018anytime}. More involved algorithms include EES~\cite{thayer2011bounded} and DPS~\cite{gilon2016dynamic}, to name a few. Recent results in bounded sub-optimal search are reported in~\cite{fickert2022new}.

Other variants to speed up the heuristic search include simultaneous usage of different heuristic functions~\cite{aine2016multi}, performing randomized heuristic search~\cite{likhachev08rstar}, etc. 

The main difference between the mentioned approaches and our work is that the former assume the heuristic function(s) to be given as the input, while in this work, we infer the heuristics from the instance of the (pathfinding) problem.

\paragraph{Machine learning for grid-based pathfinding.} 

In~\cite{takahashi2019learning,ariki2019fully}, it was proposed to use convolutional neural networks to infer the (approximate) values of the perfect cost-to-go heuristic function based on the image representation of the pathfinding instance. In~\cite{bhardwaj2017learning}, an imitation learning approach was suggested that was aimed at mimicking the behavior of the search algorithm possessing the knowledge of the perfect heuristic. In this work, the search was viewed as a sequential decision making process. Similar techniques rooted in casting the search problem as sequential decision-making were explored in~\cite{tamar2016value,panov2018grid}. More general machine learning techniques, tailored not specifically to pathfinding on grids but rather to solving combinatorial optimization problems are also known. E.g. in~\cite{https://doi.org/10.48550/arxiv.1810.10659}, a learning-based approach for solving certain NP-hard problems was presented that exploited a graph convolutional network to estimate the likelihood of whether a certain vertex of the graph is a part of the optimal solution. In~\cite{pogancic2020differentiation}, a general framework for end-to-end learning of the combination of a neural network with an arbitrary combinatorial algorithm via treating it as a piece-wise constant black-box function was proposed. Such techniques are general, although they are likely to be less efficient compared to the learnable solvers tailored specifically to the grid-based pathfinding. E.g. in the recent study of~\cite{yonetani2021path} which introduced such a tailored method, Neural A*, it was shown that the latter significantly outperform a more general technique from~\cite{pogancic2020differentiation}.  
In this work, we use Neural A* as the main baseline and show that our method consistently demonstrates better results.

\section{Background}

\subsection{Pathfinding problem}
Consider a grid, $Gr$, composed of the blocked and free cells and two distinct free grid cells, $start$ and $goal$. Being at any free cell, an agent is allowed to move to one of its cardinally- or diagonally-adjacent neighboring cells if the latter is free. The cardinal moves incur the cost of $1$ while the diagonal ones incur the cost of $\sqrt{2}$. This setting can be referred to as the 8-connected grid with non-uniform costs.

A path, $\pi(start, goal)$, is a sequence of the adjacent cells, starting with $start$ and ending with $goal$: $\pi=(c_0=start, c_1, c_2, \dots, c_n=goal)$. A path is valid \emph{iff} all the cells forming this path are free. The cost of the valid path is the sum of costs associated with the transitions between the cells comprising the path: $cost(\pi)=\sum_{i=0}^{i=n-1} cost(c_i, c_{i+1})$.

Denote a set of all valid paths connecting $start$ and $goal$ as $\Pi$. The least cost (shortest) path from $start$ to $goal$ is $\pi^* \in \Pi$, s.t. $\forall \pi \in \Pi: cost(\pi) \leq cost(\pi*)$.

The pathfinding problem is a tuple $PTask=(Gr, start, goal)$, which asks to find a \emph{valid} path from $start$ to $goal$ on $Gr$. The \emph{shortest path} is said to be the \emph{optimal solution} of $PTask$. Given a positive real number, $w > 1$, the bounded sub-optimal solution is a valid path whose cost exceeds the cost of the shortest path by no more than a factor of $w$: $cost(\pi^w) \leq w \cdot cost(\pi^*)$.

In this work, we are specifically interested in obtaining \emph{i}) valid paths; \emph{ii}) bounded sub-optimal paths. The problem of obtaining optimal solutions is out of the scope of this paper.

\subsection{A* search}
A* is a heuristic search algorithm with strong theoretical guarantees that is widely utilized to solve the pathfinding problems stated above. A* incrementally builds a search tree of nodes, where each node corresponds to a grid cell and bears the additional search-related data. This data includes the $g$-value of the node, which is the cost of the path to the node from the root of the tree. $h$-value of the node is the heuristic estimate of the cost-to-go, i.e. of the path from the current node to the goal one. The sum of $g$- and $h$-values is called the $f$-value of the node.

Nodes are generated and added to the A* search tree via the iterative \emph{expansions}. To expand a node means to generate all of its valid successors, i.e. the successors that correspond to the valid moves on a grid, to compute their $g$-values (as the sum of the $g$-value of the expanded node plus the transition cost), and to add certain successors to the tree. A successor is added to the tree only if it is not yet present in the tree or, alternatively, if the same node (i.e. the one corresponding to the same grid cell) exists, but its $g$-value is greater than the newly computed one.

A* performs expansions in the systematic fashion (starting with the $start$ node). It maintains a list of nodes that have been generated but not yet expanded. This list is typically referred to as $OPEN$, while the list of the expanded nodes is called $CLOSED$. At each iteration, a node with the minimal $f$-value is chosen from $OPEN$ for the expansion. 
A* stops when the goal node is extracted from $OPEN$. 
At this point the sought path can be reconstructed using the backpointers in the search tree. 

The performance of the algorithm, i.e. the number of iterations before termination and the guarantees on the cost of the found path, is largely dependent on the used heuristic. 

\paragraph{Heuristics} The heuristic is called perfect, denoted as $h^*$, if for every node, its value equals the true cost-to-go: $h^*(n) = cost(\pi^*(n, goal))$. The heuristic is called \emph{admissible} if it never overestimates the true cost-to-go: $h(n) \leq h^*(n)$. The heuristic is said to be \emph{consistent} or \emph{monotone} if $\forall n, n': h(n) \leq h(n') + cost(\pi^*(n, n'))$. 

A range of consistent and admissible instance-independent heuristics are known for the 8-connected grids, e.g. Chebyshev distance, Euclidean distance, or Octile distance. They all can be efficiently computed in the closed-loop form for any grid cell. Without the loss of generality, in this work, we assume that the Octile distance is used as the heuristic function.

It is known that A* with an admissible heuristic is guaranteed to find the optimal solution. Moreover, if the heuristic is \emph{consistent}, as in our case, it is not possible to find a better path to any of the expanded nodes, which infers that no node can be expanded more than once. Still, the number of such expansions can be significantly large as depicted in Fig.~\ref{fig:visual-abstract}. The reason is that the Octile distance, being an instance-independent heuristic, is unaware of the blocked cells and drives the search toward the obstacle via the low $f$-values of the nodes residing in its vicinity. 

\subsection{Weighted A* and Focal Search}
\paragraph{Weighted A*}
One of the widespread ways to trade off optimality for the computational efficiency in grid-based pathfinding is to employ a \emph{weighted heuristic}, i.e. to order nodes in $OPEN$ not by their $g+h$ values, but rather by $g+w\cdot h$ values, where $w \geq 1$. Such a modification of A*, typically referred to as WA* (Weighted A*), is known to provide bounded sup-optimal solutions w.r.t. $w$. 

\paragraph{Focal Search} Focal Search (FS)~\cite{pearl1982studies} is another technique tailored to lower the number of search iterations while providing the bound on the optimally of the resultant solution. In FS, an additional list of nodes is maintained called $FOCAL$. It is formed of the nodes residing in $OPEN$, whose $f$-values do not exceed the minimum $f$-value in $OPEN$, $f_{min}$, by a factor of $w$ (given the sub-optimality bound). 
$FOCAL$ is ordered in accordance with the secondary heuristic, $h_{FOCAL}$, which does not have to be consistent or even admissible. The node to be expanded is chosen from $FOCAL$ in accordance with ordering imposed by $h_{FOCAL}$ (and removed from $OPEN$ as well). In case $OPEN$ is updated as a result of the expansion, $FOCAL$ is modified accordingly. The stop criterion is the same as in A*. FS is guaranteed to obtain bounded sup-optimal solutions. Indeed, the number of search iterations and, thus, the computational efficiency of FS is strongly dependent on $h_{FOCAL}$.

Algorithm \ref{alg:all_in_one} shows the pseudocode of a generic heuristic search algorithm. Different colors correspond to different variants of the algorithm as explained in the caption.


\section{Method}

Recall that we are interested in two variants of the pathfinding problem. The first variant asks to find a valid path on a grid, without specifying any constraints on the cost of the path, \textsc{VP-problem}. The second variant assumes that a sub-optimality bound, $w \geq 1$, is specified and the task is to find a path whose cost does not exceed the cost of the optimal path by more than a factor of $w$, \textsc{BSP-problem}. 

The solvers that we suggest for both problems share their structure. Each of them is composed of the two building blocks. First, a deep neural network is used to process the input grid and to predict the values of the heuristic function that will be used later. Second, a heuristic search algorithm is invoked that utilizes the heuristic data from the neural network. 
The neural network used for \textsc{VP-problem} and \textsc{BSP-problem} has the same architecture; however, in each case, the output heuristic is different (as the neural network was trained using different supervision). The heuristic search algorithm is also different. For solving \textsc{VP-problem}, we utilize WA*, while for \textsc{BSP-problem} -- Focal Search (FS).

\begin{algorithm}[ht!]

\caption{A generic search algorithm. A* executes black parts only. WA* -- black and red, Focal Search -- black and purple, our variant of WA* -- black and blue.}
\label{alg:all_in_one}

\DontPrintSemicolon
\KwIn{Grid $Gr$, $start$ node, $goal$ node, heuristic function $h$, sub-optimality factor $w$, $h_{FOCAL}$ -- secondary heuristic for Focal Search}
\KwOut{path $\pi$}
$g(start) := 0$; $\forall n \neq start$ $g(n):= \infty$\\
$OPEN := \{start\}$; $CLOSED := \emptyset$\\
\While{$OPEN \neq \emptyset$}
{
    $n := GetBestNode(OPEN$\color{purple}, $FOCAL$, $h_{FOCAL}$\color{black}$)$\\
    remove $n$ from $OPEN$\color{purple} and $FOCAL$\color{black}\\
    insert $n$ into $CLOSED$\\
    \color{purple}
    \If{$f_{min}$ has changed}
    {
        update $FOCAL$
    }
    \color{black}
    \If{$n$ is $goal$}
    {
        \Return $ReconstructPath(n)$
    }
    \For{each n' in GetSuccessors(Gr,n)}
    {
        \If{$g(n') > g(n) + cost(n,n')$}
        {
            $g(n') :=  g(n) + cost(n,n')$\\
            $f(n') := g(n') + $\color{red}$w\cdot$\color{black}$h(n')$\color{blue}$/w(n')$\color{black}\\
            update or insert $n'$ in $OPEN$\\
            \color{purple}
            \If{$f(n')\leq w\cdot f_{min}$}
            {
                update or insert $n'$ in $FOCAL$\\
            }
            \color{black}
        }
    }
}
\Return \textit{path not found}\\

\end{algorithm}

\subsection{Heuristics learned}
The first type of the heuristic is the \textit{correction factor} ($cf$), which is defined as the ratio of the value of the available instance-independent heuristic
to the value of the perfect heuristic: $cf(n) = h(n)/h^*(n)$. 
We suggest plugging the predicted $cf$-values to the WA* algorithm as shown in Alg.~\ref{alg:all_in_one} (black + blue code fragments). I.e., the $f$-value of each node is computed as $f(n)=g(n)+h(n)/cf(n)$. This can be though of as running WA* that uses individual weights for different search nodes instead of a single constant weight. As there is no theoretical bound on the error of predicting $cf$-values by the neural network, the resultant search algorithm provides no guarantees on cost of the solution.


In general, predicting $cf$-values may seem similar to predicting the values of the perfect heuristic as was proposed in previous works, e.g. in~\cite{takahashi2019learning}. However, there exists a crucial difference, which is twofold. First, when learning the cost-to-go heuristic an additional technical step should be performed that transfers the range of the heuristic to the range typically employed in deep learning, e.g. $[0, 1]$. However the range of the correction factor is $[0, 1]$ by design, thus no auxiliary transformations should be made when learning. Second, the correction factor encompasses more heuristic information as it is a combination of both instance-dependent and instance-independent heuristics. As confirmed by our experiments, learning $cf$-values instead of $h^*$-values leads to a notable boost in the performance.


The second suggested heuristic is originally intended to serve as the secondary heuristic for the FS, $h_{FOCAL}$, which is tailored to solve the \textsc{BSP-problem}. Intuitively, we want from $h_{FOCAL}$ to discriminate between the nodes that are likely to yield rapid progress toward the goal and the nodes that are not. To this end, we suggest assigning (and learning to predict) a value to each grid cell that tells us how likely it is that this cell is lying on the shortest path between $start$ and $goal$. We call this value a \emph{path probability}, $pp$-value, and, by design, its range is also within $[0, 1]$. Learning to correctly predict $pp-values$ may be thought of as attempting to learn to solve the pathfinding queries directly. I.e., if we were able to obtain such predictions for each problem instance where $pp$-values of $1$ were assigned to the cells lying on the shortest path, while the other cells were assigned $pp$-values of $0$, then we would not need to run the search algorithm at all. However, in practice, this is not realistic, thus we use the predicted $pp$-values as $h_{FOCAL}$ values in the FS. 


\begin{figure*}[ht!]
    \centering
    \includegraphics[width=\linewidth]{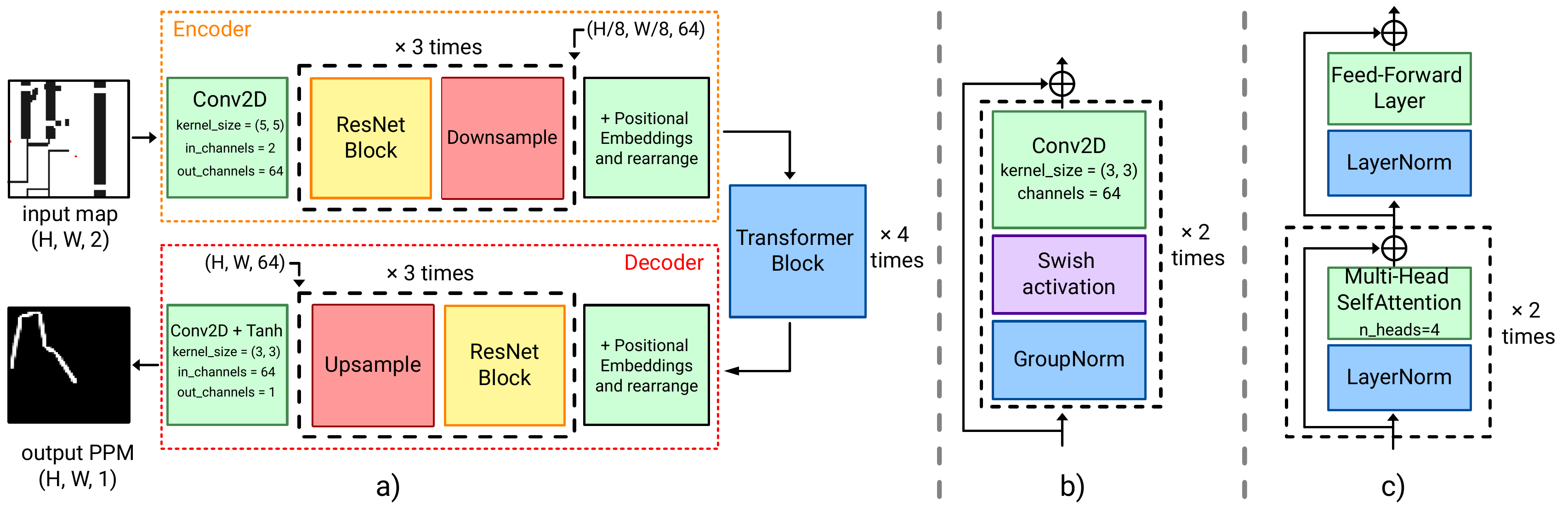}
    \caption{Overview of the neural network architecture. a) Design of the whole model. CNN-encoder is used to produce local features which are further fed into the transformer blocks to catch the long-range dependencies between the features. The resulting representation is passed through the CNN-decoder to produce output values. b) Architecture of the ResNet block. c) Architecture of the Transformer block.}
    \label{fig:gen_arch}
\end{figure*}

\subsection{Learning supervision}
An evident approach to learn the suggested heuristics is to create a rich dataset of pathfinding instances with the annotated ground-truth $cf$-values and $pp$-values and to learn the neural network to minimize the error between its predictions and ground-truth values. Thus, no search/planning is happening at the training phase. Using the techniques introduced in~\cite{pogancic2020differentiation,yonetani2021path}, one might consider another option of learning, i.e. to include the search algorithm in the learning pipeline and to back-propagate the search error through it. This option is especially useful when it is hard or impossible to create ground-truth samples (like in planning on images when the cost of the transition from pixel to pixel in unknown). We have experimented (see Appendix~\ref{appendix:training_modes} for details) with both types of learning and found that for the considered setting, the first option is preferable for the following reasons. First, it is not a problem to create ground-truth samples for $cf$- and $pp$-values (technical details on this will follow shortly). Second, learning without differentiable planner is much faster (up to 4x in our setup). Third and not least of all, our experiments showed that learning the suggested model with the supervision from the ground-truth values leads to a consistently better performance.

To create ground-truth $cf$-values, we utilized uninformed search that starts in the goal location on a map and computes true distances to the goal from any cell (which are straightforwardly converted to the $cf$-values). Creating the ground-truth samples of $pp$-values (we refer to such samples as path probability maps, or PPMs) is more involved. Recall that in PPMs, we need to have values of $1$ for the cells lying on the shortest path while all other cells should have smaller values. However, numerous shortest paths on 8-connected grids might exist. Moreover, many of them may differ only by the ordering of cardinal and diagonal moves. To this end, we ran Theta*~\cite{nash2007}, an any-angle search algorithm that can be though of as A* with online path smoothing. Theta* paths are formed of the way points (cells) located at the corners of the obstacles and cells that lie on the straight-line segments connecting the way points. In the resultant PPM, we assign the values of $1$ for such paths. For all other cells, we computed the value that tells us how close the cost of the path through cell $n$ to the one of Theta* is: $cost(\pi(s,g))/(cost(\pi(s,n)+cost(\pi(n,g))$. If this value was greater than or equal to $0.95$, we used it as the $pp$-value; if not, the $pp$-value was set to $0$. As a result, we obtain PPMs that contain paths from $start$ to $goal$ and narrow tunnels (with lower $pp$-values) around them (see Fig.~\ref{fig:visual-abstract}).

\subsection{Neural network architecture}
The neural network for learning $cf$-values and $pp$-values has the same architecture, however the input is slightly different. For $pp$-values the input contains the grid, represented as a binary image 
and the start-goal matrix, which has the same size as the grid and contains the values of $1$ only for start and goal, while all other pixels are zeroes. For $cf$-values this matrix contains only one non-zero element -- the goal one.

\begin{figure*}[ht!]
    \centering
    \includegraphics[width=\linewidth]{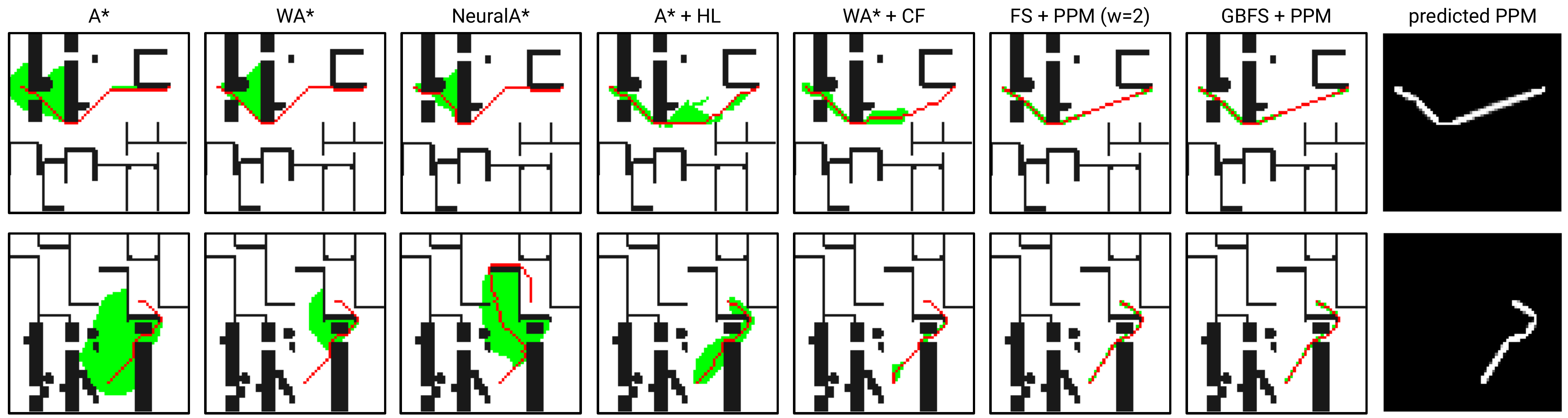}
    \caption{Several examples of the pathfinding results. The expanded nodes are shown in green, the path -- in red. The last column shows the predicted PPMs}
    \label{fig:solved_instance_examples}
\end{figure*}

The architecture has three main blocks (see Fig.~\ref{fig:gen_arch}): convolutional encoder, spatial transformer and convolutional decoder. Convolutional encoder utilizes the well-known ResNet blocks~\cite{He_2016_CVPR} and is aimed at extracting the local features of the pathfidning instance such as corners of the obstacles, narrow passages etc. Transformer leverages the mechanism of self-attention~\cite{vaswani2017attention} to establish the global relations between these features (how important is one feature w.r.t. the other). An example may be, how is it important that there is a narrow passage in between start and goal. Transformers were originally suggested for text sequences that lack 2D structure. However, in the considered case this structure is important. To this end, we utilize the positional embedding technique from Visual Transformers~\cite{dosovitskiy2021an, 9716741}. This technique re-arranges 2D feature maps into vectors (before the transformer block) and vice-versa (afterwards), while preserving the spatial structure. Finally, the transformed feature maps of the pathfinding instances are processed by the convolutional decoder, which provides the final output.


\section{Training and Evaluation}

\subsection{Dataset}

We have adopted the TMP (Tiled Motion Planning) dataset that was used in~\cite{yonetani2021path} for empirical evaluation. This dataset is a modification of the MP dataset used in~\cite{bhardwaj2017learning}. The latter consists of maps with various challenging topologies, such as bugtraps, gaps, etc. Each map in the TMP dataset is composed of the four MP maps. In total, $4,000$ maps of size $64\times64$ were present in TMP.

We further increased the size of the dataset to $64,000$ maps via the augmentation by mirroring and rotating each of the four parts of the TMP maps. Examples are shown in Fig.~\ref{fig:solved_instance_examples}. For each map, we generated 10 problem instances. Goal location was chosen randomly, while start one was chosen randomly out of the $1/3$ of the reachable nodes that have the highest cost of the path from the goal. Overall we generated $640,000$ instances. They were divided in the proportion of 8:1:1 for train, validation, and test subsets in such a way that all augmented versions of the same map were presented only in one of the subsets. Similarly to~\cite{takahashi2019learning} we have excluded from the test part of the dataset the instances that are extremely easy to solve, more formally -- the ones that have hardness less than $1.05$. Here hardness is defined as  $cost(\pi^*(start, goal))/h(start)$, where $h$ is the conventional cost-to-go heuristic. The closer this value is to $1.0$ the easier the instance is, meaning that there is almost no need to bypass the obstacles and the path resembles a straight line in the free space.

\subsection{Planners}
We denote the planners proposed in this work as WA*+CF (Weighted A* with the correction factor), FS+PPM (Focal Search with Path Probabilty Map). We also evaluated a combination of the Greedy Best First Search with PPM, denoted as GBPS+PPM. This planner greedily selects nodes by their $pp$-values (preferring the ones with the smaller $f$-values to break ties) and, thus, does not guarantee bounded sup-optimality of the solution. \footnote{The source-code of our planners is publicly available at \url{https://www.github.com/AIRI-Institute/TransPath}.}

The baselines we compare against include both standard heurstic search algorithms, A* and WA*, as well as the learnable ones. The latter are represented by the two planners. The first one is Neural A*~\cite{yonetani2021path}, the state-of-the-art planner that was shown to notably outperform a range of competitors.
The second is the planner from ~\cite{takahashi2019learning}, which predicts the perfect cost-to-go heuristic was and use in for A*. We denote this approach as A*+HL. We used the official code of Neural A* and modified it to handle non-uniform move costs. Moreover, we employed our neural network model in Neural A* to provide a fairer comparison (the performance of Neural A* with the original neural network was significantly worse according to our preliminary experiments). Similarly we used our neural network for predicting cost-to-go heuristic in A*+HL. For bounded sub-optimal planners, i.e. WA* and FS+PPM, we set the sub-otimality factor as $w=2$. 

\subsection{Training setup}
To train the neural networks predicting $cf$-values, $pp$-values and cost-to-go estimates (for A*+HL) we use the same setup. We train each model using Adam optimizer~\cite{kingma2014adam} for $35$ epochs with a batch size of $512$ and OneCycleLR learning rate scheduler~\cite{DBLP:journals/corr/abs-1708-07120} at a maximum learning rate of $4 \times 10^{-4}$. We use $L_2$ loss for $cf$-values, $pp$-values and $L_1$ loss for the cost-to-go estimates following~\cite{takahashi2019learning}. It took us $3.5$ hours to train each model on NVIDIA A100 80GB GPU.


We trained Neural A* on our training data with the same training setup as in the original work. It took 16 hours to learn the model on our hardware, 4 times more compared to $cf$-/$pp$-values. This is expected, as Neural A* is trained with the differentiable A* in the loop.

\begin{table}[ht!]
    \centering
    \resizebox{.99\columnwidth}{!}{%
    \begin{tabular}{c|ccc}
    &Optimal Found &Cost &Expansions\\
    &Ratio ($\%$) $\uparrow$ & Ratio ($\%$) $\downarrow$ & Ratio ($\%$) $\downarrow$\\
    \hline
    A*& 100 & 100 & 100\\
    WA*& 40.66 & 103.52 $\pm$4.85 & 44.43 $\pm$25.92\\
    Neural A*& 29.82&104.90 $\pm$6.56&52.30 $\pm$30.47\\
    A*+HL& 79.11&100.27 $\pm$0.62&80.50 $\pm$74.40\\
    WA*+CF&\textbf{85.40}&100.25 $\pm$1.13&36.98 $\pm$21.18\\
    FS+PPM&82.97&\textbf{100.24 $\pm$0.74}&26.36 $\pm$21.08\\
    GBFS+PPM&83.02&100.25 $\pm$0.90&\textbf{23.60 $\pm$18.34}\\

    \end{tabular}%
    }
    \caption{Experimental results. Values before $\pm$ indicate the average, while after $\pm$ -- the standard deviation.}
    \label{tab:exp}
\end{table}

\subsection{Results}

We were primarily interested in the following performance metrics: Expansions Ratio -- the ratio of the number of expansions performed by the planner to the number of A* expansions; Cost Ratio -- the same ratio but for the solution cost; Optimal Found Ratio -- the ratio of instances optimally solved by the planner.

Table~\ref{tab:exp} shows the average values and standard error of these indicators for the test dataset. Clearly, all the learning-based planners are able to generalize to the unseen instances solving them near-optimally while reducing the search effort. In terms of Cost Ratio the best results are demonstrated by FS+PPM, while the other our planner, WA*+CF, turned out to outperform the competitors in terms of number of the instances solved optimally. The number of reduced expansions varied significantly for all algorithms (see the third column after the $\pm$ sign) and, evidently, in certain cases one of the learnable planners, i.e. A*+HL, managed to expand more nodes than A*. Still, the techniques suggested in this work, in particular predicting $pp$-values in combination with FS and GBFS, managed to reduce the number of the expansions significantly (up to 4 times approximately) in numerous cases, as the average value of the Expansions Ratio tells.

Fig.~\ref{fig:solved_instance_examples} shows several instances from the dataset with the solutions obtained by the evaluated algorithms and the nodes they expand. The techniques suggested in this work clearly outperform the competitors. 

More details can be found in Appendix~\ref{appendix:det_results}.

\paragraph{Runtime breakdown} We measured the runtime of the compared methods, though it is heavily dependent on the implementation and the hardware. E.g. Neural A* is fully implemented in Python, while our planners feature both Python for neural networks' machinery and C++ for the search itself. Thus, it is not correct to directly compare their runtimes. To this end we do not report the runtime of Neural A*. As for the other methods (implemented solely by us) the breakdown of their runtimes are as follows. Prediction step for batch size of $64$ and the native torch float32 type required ~9.5ms on Tesla A100 GPU (and ~40ms on GTX 1660S). The average CPU time required for further solving this batch of $64$ tasks: A* -- 155ms, WA* -- 77ms, WA*+CF -- 60ms, A*+HL -- 96ms, FS+PPM -- 37ms, GBFS+PPM -- 31ms.

\begin{table}[t!]
    \centering
    \resizebox{.99\columnwidth}{!}{%
    \begin{tabular}{c|cc}
    FS+PPM ($w=4$) & w/ Trans & w/o Trans \\
    \hline
Optimal Found Ratio (\%) & 85.22	& 61.74\\
Average Cost Ratio (\%) & 100.31 $\pm$ 1.58	&	101.12 $\pm$ 2.19 \\
Average Expansions Ratio (\%) & 16.06 $\pm$ 11.57 	&	19.65 $\pm$ 17.03		\\
MSE $\times 10^{-3}$ & \textbf{3.2} & 5.3   \\

    \end{tabular}%
    }
    \caption{Ablation study results. Values before $\pm$ indicate the average, while after $\pm$ -- the standard deviation.}
    \label{tab:transblocks}
\end{table}

\begin{figure}[t]
    \centering
    \includegraphics[width=0.75\columnwidth]{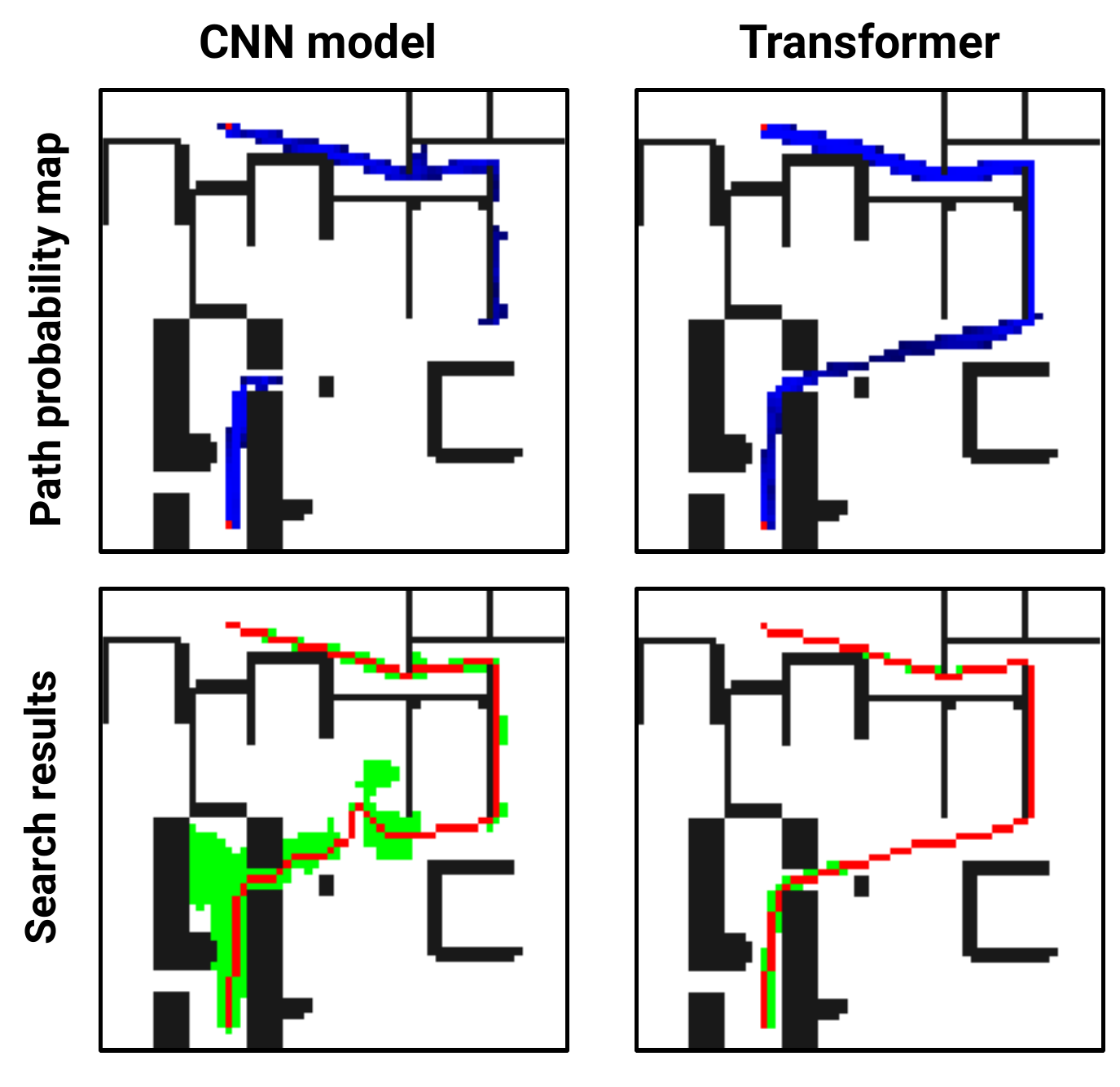}
    \caption{An example showing the difference between the CNN (only) model and the one with the Transformer block.}
    \label{fig:cnn_trans}
\end{figure}

\subsection{Evaluation on out-of-the-distribution dataset} 
We additionally evaluated all the considered planners on a range of maps that differ significantly in topology and size from the maps of TMP dataset. More details on this experiment can be found in Appendix~\ref{appendix:out_of_distribution}. Overall, the results are similar to the ones reported above -- our methods outperform the competitors.

\subsection{Ablation study}
To demonstrate the importance of using the Transformer block in the neural network we created a version of the latter that omits this block and is composed only of the convolutional layers (CNN model). We trained this neural network similarly to the baseline model. To compare them we selected tasks with the hardness exceeding $1.5$ as we hypothesize that utilizing transformer is especially useful for non-trivial instances. Quantitative results are presented in Table~\ref{tab:transblocks} while qualitative in Fig.~\ref{fig:cnn_trans} (we demonstrate the results for FS+PPM for the sake of space, as the results for WA*+CF are similar).

Clearly the usage of transformers notably increases the performance across all of the considered metrics, as Table~\ref{tab:transblocks} tells. The last row reports the mean squared error (MSE) between the predictions of the neural network and the ground-truth values.

As Fig.~\ref{fig:cnn_trans} shows, transformer allows to capture the long-range dependencies between the regions of interest on the map and, consequently, to form a complex and accurate PPM, which aids the search algorithm substantially. The PPM of the CNN model is, however, fragmented and, as a result, a less natural-looking path is produced while the number of expansions is higher.

\section{Conclusion}

In this work we explored how state-of-the-art deep learning technique may aid heuristic search planners in solving grid-based pathfinding problems. We have suggested to utilize a deep neural network composed of both convolutional and attention layers to predict several novel heuristics that can be used in combination with Weighed A* and Focal Search. We showed empirically that the suggested planners were able to successfully solve challenging problems and outperform the competitors that include both traditional heuristic search techniques as well as the state-of-the-art learnable approaches.

The avenues for future research include, but not limited to: planning in 3D, planning with kinodynamic constraints (including sample-based planning) etc.

\bibliography{main.bib}

\begin{thebibliography}{32}
\providecommand{\natexlab}[1]{#1}

\bibitem[{Aine et~al.(2016)Aine, Swaminathan, Narayanan, Hwang, and
  Likhachev}]{aine2016multi}
Aine, S.; Swaminathan, S.; Narayanan, V.; Hwang, V.; and Likhachev, M. 2016.
\newblock Multi-heuristic a.
\newblock \emph{The International Journal of Robotics Research}, 35(1-3):
  224--243.

\bibitem[{Ariki and Narihira(2019)}]{ariki2019fully}
Ariki, Y.; and Narihira, T. 2019.
\newblock Fully convolutional search heuristic learning for rapid path
  planners.
\newblock \emph{arXiv preprint arXiv:1908.03343}.

\bibitem[{Bhardwaj, Choudhury, and Scherer(2017)}]{bhardwaj2017learning}
Bhardwaj, M.; Choudhury, S.; and Scherer, S. 2017.
\newblock Learning heuristic search via imitation.
\newblock In \emph{Proceedings of the 1st Conference on Robot Learning ({CoRL}
  2017)}, 271--280.

\bibitem[{Cohen et~al.(2018)Cohen, Greco, Ma, Hern{\'a}ndez, Felner, Kumar, and
  Koenig}]{cohen2018anytime}
Cohen, L.; Greco, M.; Ma, H.; Hern{\'a}ndez, C.; Felner, A.; Kumar, T.~S.; and
  Koenig, S. 2018.
\newblock Anytime Focal Search with Applications.
\newblock In \emph{Proceedings of the 27th International Joint Conference on
  Artificial Intelligence ({IJCAI} 2018)}, 1434--1441.

\bibitem[{Dosovitskiy et~al.(2021)Dosovitskiy, Beyer, Kolesnikov, Weissenborn,
  Zhai, Unterthiner, Dehghani, Minderer, Heigold, Gelly, Uszkoreit, and
  Houlsby}]{dosovitskiy2021an}
Dosovitskiy, A.; Beyer, L.; Kolesnikov, A.; Weissenborn, D.; Zhai, X.;
  Unterthiner, T.; Dehghani, M.; Minderer, M.; Heigold, G.; Gelly, S.;
  Uszkoreit, J.; and Houlsby, N. 2021.
\newblock An Image is Worth 16x16 Words: Transformers for Image Recognition at
  Scale.
\newblock In \emph{Proceedings of the 9th International Conference on Learning
  Representations ({ICLR} 2021)}.

\bibitem[{Fickert, Gu, and Ruml(2022)}]{fickert2022new}
Fickert, M.; Gu, T.; and Ruml, W. 2022.
\newblock New Results in Bounded-Suboptimal Search.
\newblock In \emph{Proceedings of the 36th {AAAI} Conference on Artificial
  Intelligence ({AAAI} 2022)}, 10166--10173.

\bibitem[{Gilon, Felner, and Stern(2016)}]{gilon2016dynamic}
Gilon, D.; Felner, A.; and Stern, R. 2016.
\newblock Dynamic potential search -- a new bounded suboptimal search.
\newblock In \emph{Proceedings of the 9th Annual Symposium on Combinatorial
  Search ({SOCS} 2016)}, 36--44.

\bibitem[{Han et~al.(2022)Han, Wang, Chen, Chen, Guo, Liu, Tang, Xiao, Xu, Xu,
  Yang, Zhang, and Tao}]{9716741}
Han, K.; Wang, Y.; Chen, H.; Chen, X.; Guo, J.; Liu, Z.; Tang, Y.; Xiao, A.;
  Xu, C.; Xu, Y.; Yang, Z.; Zhang, Y.; and Tao, D. 2022.
\newblock A Survey on Vision Transformer.
\newblock \emph{IEEE Transactions on Pattern Analysis and Machine
  Intelligence}, 1--1.

\bibitem[{Hansen and Zhou(2007)}]{hansen2007anytime}
Hansen, E.~A.; and Zhou, R. 2007.
\newblock Anytime heuristic search.
\newblock \emph{Journal of Artificial Intelligence Research}, 28: 267--297.

\bibitem[{Hart, Nilsson, and Raphael(1968)}]{hart1968formal}
Hart, P.~E.; Nilsson, N.~J.; and Raphael, B. 1968.
\newblock A formal basis for the heuristic determination of minimum cost paths.
\newblock \emph{IEEE transactions on Systems Science and Cybernetics}, 4(2):
  100--107.

\bibitem[{He et~al.(2016)He, Zhang, Ren, and Sun}]{He_2016_CVPR}
He, K.; Zhang, X.; Ren, S.; and Sun, J. 2016.
\newblock Deep Residual Learning for Image Recognition.
\newblock In \emph{Proceedings of the IEEE Conference on Computer Vision and
  Pattern Recognition (CVPR)}.

\bibitem[{Janner et~al.(2022)Janner, Du, Tenenbaum, and Levine}]{Janner2022}
Janner, M.; Du, Y.; Tenenbaum, J.~B.; and Levine, S. 2022.
\newblock {Planning with Diffusion for Flexible Behavior Synthesis}.
\newblock In \emph{ICML}.

\bibitem[{Kingma and Ba(2014)}]{kingma2014adam}
Kingma, D.~P.; and Ba, J. 2014.
\newblock Adam: A method for stochastic optimization.
\newblock \emph{arXiv preprint arXiv:1412.6980}.

\bibitem[{Li, Chen, and
  Koltun(2018)}]{https://doi.org/10.48550/arxiv.1810.10659}
Li, Z.; Chen, Q.; and Koltun, V. 2018.
\newblock Combinatorial Optimization with Graph Convolutional Networks and
  Guided Tree Search.

\bibitem[{Likhachev, Gordon, and Thrun(2003)}]{likhachev2004ara}
Likhachev, M.; Gordon, G.~J.; and Thrun, S. 2003.
\newblock {ARA*} : Anytime {A*} with Provable Bounds on Sub-Optimality.
\newblock In Thrun, S.; Saul, L.~K.; and Sch\"{o}lkopf, B., eds.,
  \emph{Advances in Neural Information Processing Systems 16 ({NIPS} 2003)},
  767--774. MIT Press.

\bibitem[{Likhachev and Stentz(2008)}]{likhachev08rstar}
Likhachev, M.; and Stentz, A. 2008.
\newblock R* Search.
\newblock In \emph{Proceedings of the 23rd {AAAI} Conference on Artificial
  Intelligence ({AAAI} 2008)}, 344--350. {AAAI} Press.

\bibitem[{Nash et~al.(2007)Nash, Daniel, Koenig, and Felner}]{nash2007}
Nash, A.; Daniel, K.; Koenig, S.; and Felner, A. 2007.
\newblock Theta*: Any-Angle Path Planning on Grids.
\newblock In \emph{Proceedings of The 22nd AAAI Conference on Artificial
  Intelligence ({AAAI} 2007)}, 1177--1183.

\bibitem[{Panov, Yakovlev, and Suvorov(2018)}]{panov2018grid}
Panov, A.~I.; Yakovlev, K.~S.; and Suvorov, R. 2018.
\newblock Grid path planning with deep reinforcement learning: Preliminary
  results.
\newblock \emph{Procedia computer science}, 123: 347--353.

\bibitem[{Pearl and Kim(1982)}]{pearl1982studies}
Pearl, J.; and Kim, J.~H. 1982.
\newblock Studies in semi-admissible heuristics.
\newblock \emph{IEEE transactions on pattern analysis and machine
  intelligence}, (4): 392--399.

\bibitem[{Pogan{\v{c}}i{\'c} et~al.(2020)Pogan{\v{c}}i{\'c}, Paulus, Musil,
  Martius, and Rolinek}]{pogancic2020differentiation}
Pogan{\v{c}}i{\'c}, M.~V.; Paulus, A.; Musil, V.; Martius, G.; and Rolinek, M.
  2020.
\newblock Differentiation of blackbox combinatorial solvers.
\newblock In \emph{Proceedings of the 8th International Conference on Learning
  Representations ({ICLR} 2020)}.

\bibitem[{Pohl(1970)}]{pohl1970heuristic}
Pohl, I. 1970.
\newblock Heuristic search viewed as path finding in a graph.
\newblock \emph{Artificial intelligence}, 1(3-4): 193--204.

\bibitem[{Ramachandran, Zoph, and Le(2017)}]{DBLP:journals/corr/abs-1710-05941}
Ramachandran, P.; Zoph, B.; and Le, Q.~V. 2017.
\newblock Searching for Activation Functions.
\newblock \emph{CoRR}, abs/1710.05941.

\bibitem[{Rivera et~al.(2020)Rivera, Hern{\'a}ndez, Hormaz{\'a}bal, and
  Baier}]{rivera20202}
Rivera, N.; Hern{\'a}ndez, C.; Hormaz{\'a}bal, N.; and Baier, J.~A. 2020.
\newblock The 2\^{} k Neighborhoods for Grid Path Planning.
\newblock \emph{Journal of Artificial Intelligence Research}, 67: 81--113.

\bibitem[{Smith and Topin(2017)}]{DBLP:journals/corr/abs-1708-07120}
Smith, L.~N.; and Topin, N. 2017.
\newblock Super-Convergence: Very Fast Training of Residual Networks Using
  Large Learning Rates.
\newblock \emph{CoRR}, abs/1708.07120.

\bibitem[{Speck et~al.(2021)Speck, Biedenkapp, Hutter, Mattm{\"{u}}ller, and
  Lindauer}]{Speck2020}
Speck, D.; Biedenkapp, A.; Hutter, F.; Mattm{\"{u}}ller, R.; and Lindauer, M.
  2021.
\newblock {Learning Heuristic Selection with Dynamic Algorithm Configuration}.
\newblock In \emph{Proceedings of the Thirty-First International Conference on
  Automated Planning and Scheduling}, volume~31, 597--605.

\bibitem[{Sturtevant(2012)}]{sturtevant2012benchmarks}
Sturtevant, N.~R. 2012.
\newblock Benchmarks for Grid-Based Pathfinding.
\newblock \emph{IEEE Transactions on Computational Intelligence and AI in
  Games}, 4(2): 144--148.

\bibitem[{Takahashi et~al.(2019)Takahashi, Sun, Tian, and
  Wang}]{takahashi2019learning}
Takahashi, T.; Sun, H.; Tian, D.; and Wang, Y. 2019.
\newblock Learning heuristic functions for mobile robot path planning using
  deep neural networks.
\newblock In \emph{Proceedings of the 29th International Conference on
  Automated Planning and Scheduling ({ICAPS} 2019)}, 764--772.

\bibitem[{Tamar et~al.(2016)Tamar, Wu, Thomas, Levine, and
  Abbeel}]{tamar2016value}
Tamar, A.; Wu, Y.; Thomas, G.; Levine, S.; and Abbeel, P. 2016.
\newblock Value iteration networks.

\bibitem[{Tan and Le(2019)}]{DBLP:journals/corr/abs-1905-11946}
Tan, M.; and Le, Q.~V. 2019.
\newblock EfficientNet: Rethinking Model Scaling for Convolutional Neural
  Networks.
\newblock \emph{CoRR}, abs/1905.11946.

\bibitem[{Thayer and Ruml(2011)}]{thayer2011bounded}
Thayer, J.~T.; and Ruml, W. 2011.
\newblock Bounded suboptimal search: A direct approach using inadmissible
  estimates.
\newblock In \emph{Proceedings of the 22nd International Joint Conference on
  Artificial Intelligence ({IJCAI}) 2011)}, 674--679.

\bibitem[{Vaswani et~al.(2017)Vaswani, Shazeer, Parmar, Uszkoreit, Jones,
  Gomez, Kaiser, and Polosukhin}]{vaswani2017attention}
Vaswani, A.; Shazeer, N.; Parmar, N.; Uszkoreit, J.; Jones, L.; Gomez, A.~N.;
  Kaiser, {\L}.; and Polosukhin, I. 2017.
\newblock Attention is all you need.
\newblock \emph{Advances in neural information processing systems}, 30.

\bibitem[{Yonetani et~al.(2021)Yonetani, Taniai, Barekatain, Nishimura, and
  Kanezaki}]{yonetani2021path}
Yonetani, R.; Taniai, T.; Barekatain, M.; Nishimura, M.; and Kanezaki, A. 2021.
\newblock Path planning using neural {A*} search.
\newblock In \emph{Proceedings of the 38th International Conference on Machine
  Learning (ICML 2021)}, 12029--12039.

\end{thebibliography}

\newpage

\clearpage

\setcounter{secnumdepth}{1}
\begin{appendices}
\section{Different Modes of Training} \label{appendix:training_modes}

We have examined whether the training approach used for NeuralA*~\cite{yonetani2021path}, is beneficial for our problem setup. NeuralA* is a combination of the neural network (CNN-based encoder-decoder) and the differentiable A* search. The loss function used for training NeuralA* is the mean absolute error between the map of the expanded nodes, i.e. the matrix whose elements are equal to 1 if the corresponding node was expanded and 0 otherwise, and the matrix-encoded ground truth path, i.e. the matrix which elements are 1 if the corresponding grid cells lie on the shortest path. This way the error, backpropagated through both the differentiable A* and the neural network, is, essentially, the search error.

We have modified the code of NeuralA* for 8-connected non-uniform cost grids (originally NeuralA* was tailored to uniform cost grids) and made it learn the correction factor ($cf$-values). We have also substituted the less advanced neural network backbone of NeuralA* with our transformer-based model. We trained it with the original training setup used in~\cite{yonetani2021path}.

Thus, we ended up with the two models that differed only in the mode of training. One model was trained to minimize the error between the predicted $cf$-values and the ground-truth ones (direct supervision), and the other was trained to minimize the search error that arise from using the predicted $cf$-values in differentiable A*.

After training we evaluated both models on the instances of the test split of our dataset. Table~\ref{tab:dif_vs_super} shows the results. As one can see, the model trained by direct supervision (i.e. by utilizing the ground-truth $cf$-values) was consistently better.

\begin{table}[ht!]
    \centering
    \resizebox{.99\columnwidth}{!}{%
    \begin{tabular}{c|cc}
    WA* + CF & Learning gt $cf$-values & Differentiable A* \\
    \hline
Optimal Found Ratio (\%) & \textbf{85.40}	& 29.15 \\
Average Cost Ratio (\%) & \textbf{100.25 $\pm$ 1.13}	&	103.04 $\pm$ 3.71 \\
Average Expansions Ratio (\%) & \textbf{36.98 $\pm$ 21.18} 	&	48.47 $\pm$ 28.50		\\

    \end{tabular}%
    }
    \caption{Quantitative comparison of different modes of training. Values before $\pm$ indicate the average, while after $\pm$ -- the standard deviation.}
    \label{tab:dif_vs_super}
\end{table}

Figure~\ref{fig:cnn_trans} depicts one example of the learned $cf$-values and the corresponding search results for both models. NeuralA* seems to learn a somewhat different representation of the correction factor, which is, indeed, beneficial for search. However with direct supervision the search result is, evidently, better, i.e. the number of the expanded nodes is smaller. The other examples that we have manually examined confirm this observation.

Overall, based on the reported results we have chosen the direct supervision to be the mode of training used in our work.

\begin{figure}[h!]
    \centering
    \includegraphics[width=0.9\columnwidth]{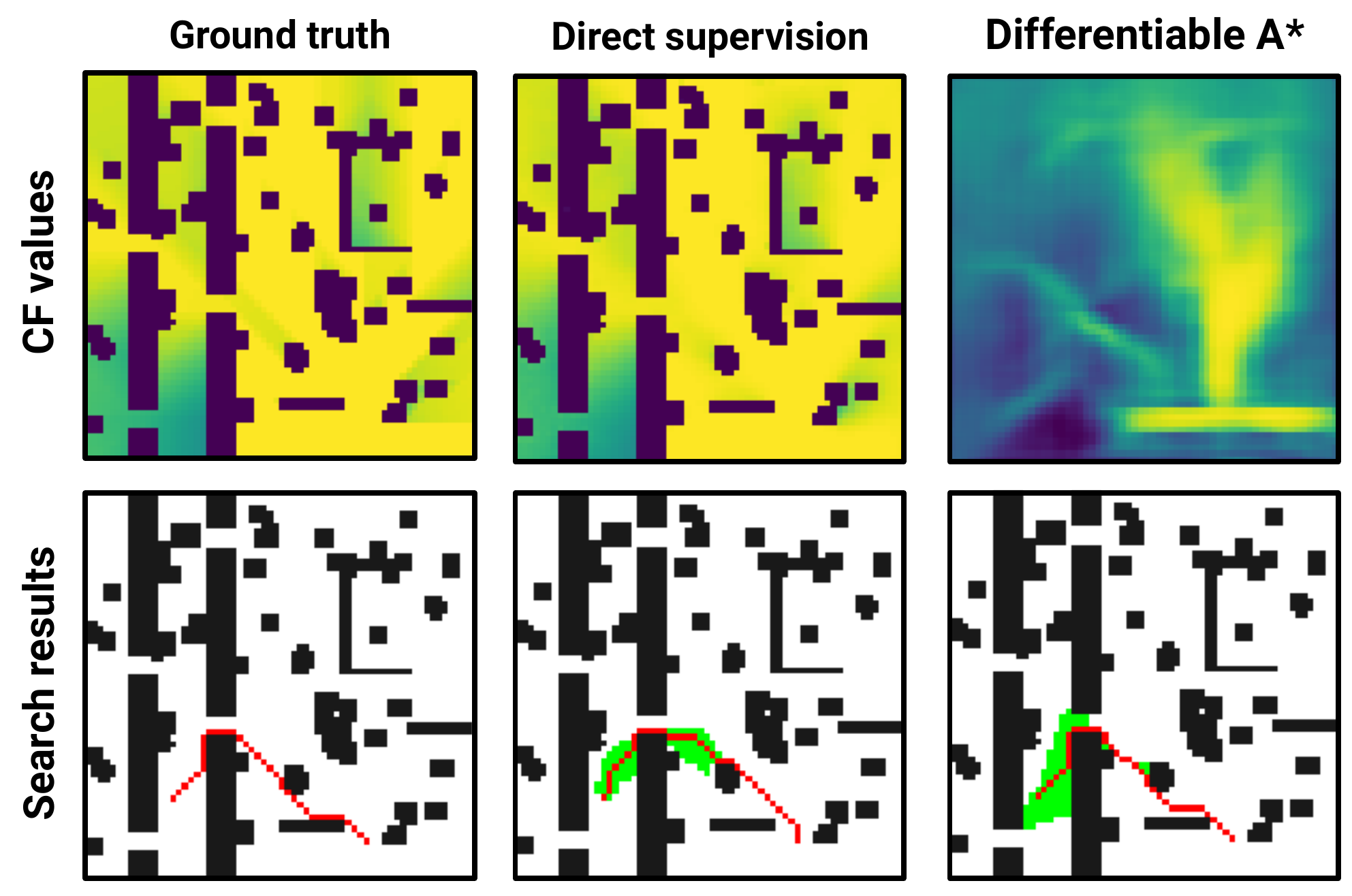}
    \caption{An example of the $cf$-values (and the corresponding search results) learned via the direct supervision and with the differentiable A* in the loop.}
    \label{fig:cnn_trans}
\end{figure}

\section{More Details On The Empirical Results}
\label{appendix:det_results}

In the main body of the paper we reported the average values of the metrics we were interested in and the standard error across all the instances of our test dataset. Here we present more detailed results, i.e. the box-n-whisker plots for the instances grouped together based on their hardness. Recall, that  hardness is defined as $cost(\pi^*(start, goal))/h(start)$, where $h$ is the conventional cost-to-go heuristic (Octile distance in our case) and $\pi^*$ is the shortest path from $start$ to $goal$. The closer this value is to $1.0$ the easier the instance is, meaning that there is almost no need to bypass the obstacles and the path resembles a straight line in the free space. Recall also, that the minimal hardness for our test instances is $1.05$.

\begin{figure*}[h!]
     \centering
     \begin{subfigure}[b]{0.44\textwidth}
         \centering
         \includegraphics[width=\textwidth]{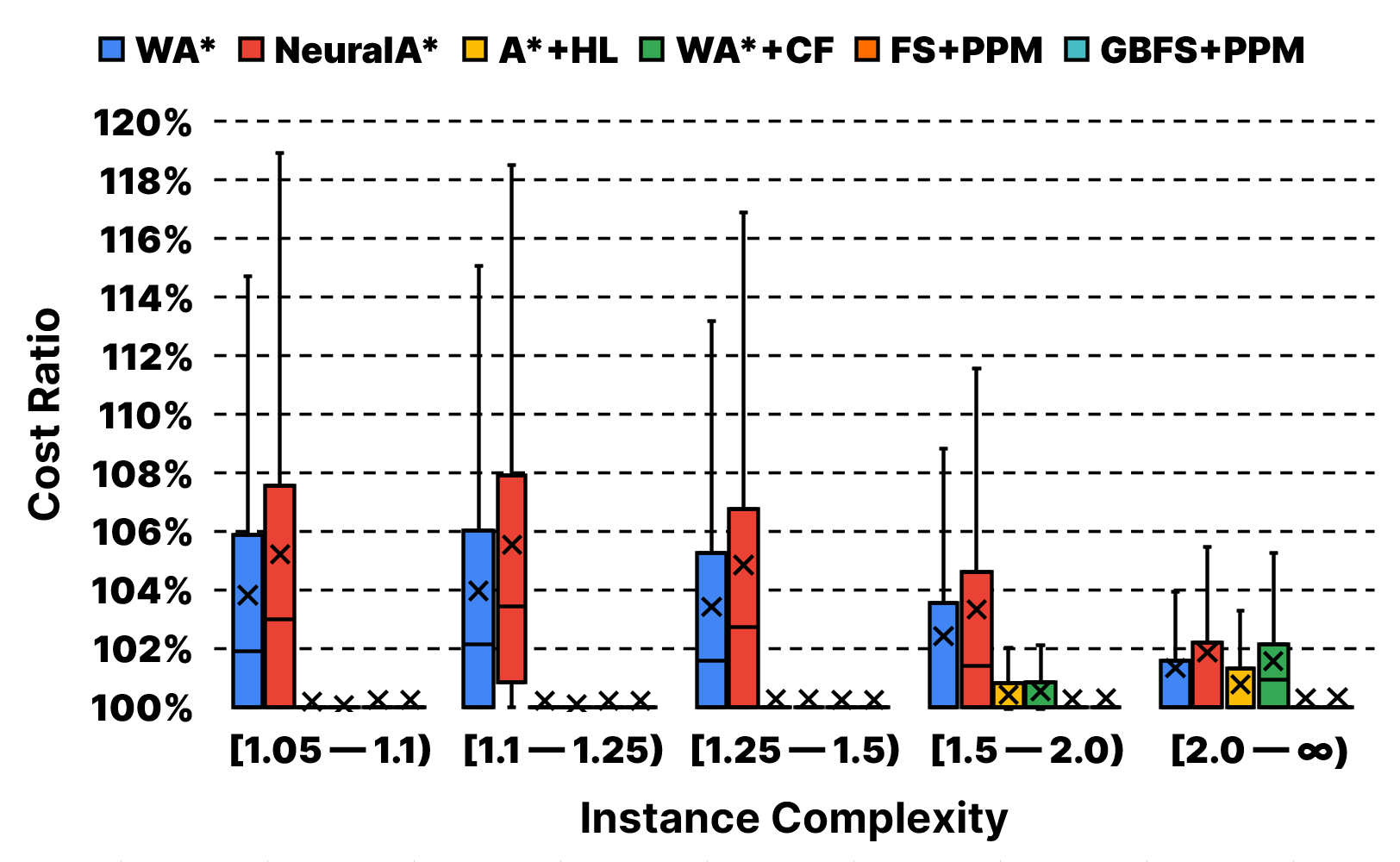}
         \caption{Box-and-whiskers plots of the cost ratio.}
     \end{subfigure}
     \begin{subfigure}[b]{0.44\textwidth}
         \centering
         \includegraphics[width=\textwidth]{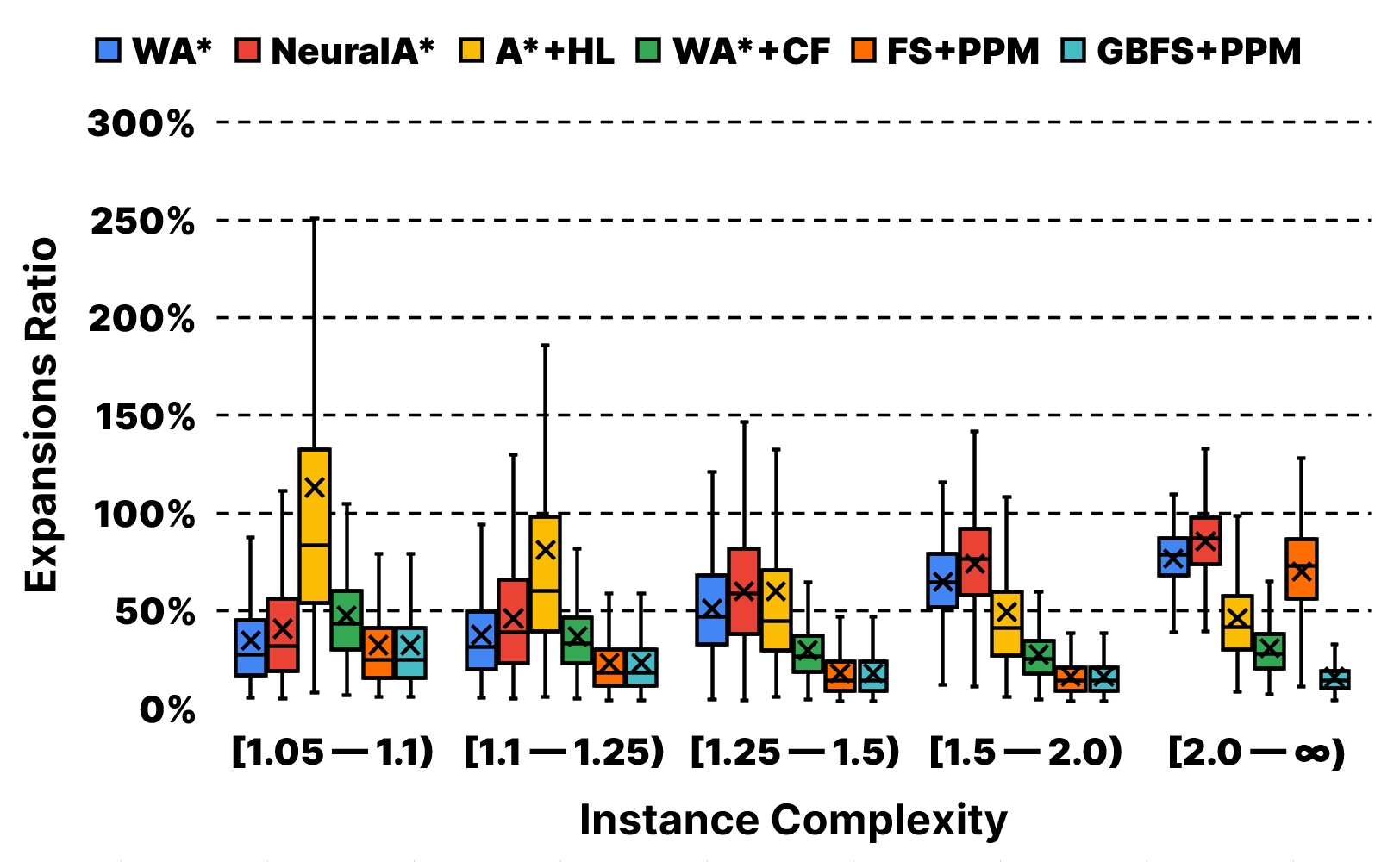}
         \caption{Box-and-whiskers plots of the expansions ratio.}
     \end{subfigure}
        \caption{Cost and expansions ratios w.r.t. the hardness of the test instances.}
        \label{fig:det_res}
\end{figure*}

Fig.~\ref{fig:det_res} shows the box-n-whisker plots. As one can note, the cost ratio of WA* and NeuralA* decreases when the instances get harder, however it is not the case for the other planners. For very hard instances (with hardness exceeding 2) WA*, NeuralA*, A*+HL, WA*+CF demonstrate similar results. FS+PPM and GBFS+PPM are indeed the ultimate winners in terms of cost ratio. More importantly, their performance does not seem to be tied to the hardness of the pathfinding instances they are facing.

\begin{figure*}[th!]
    \centering
    \includegraphics[width=0.9\linewidth]{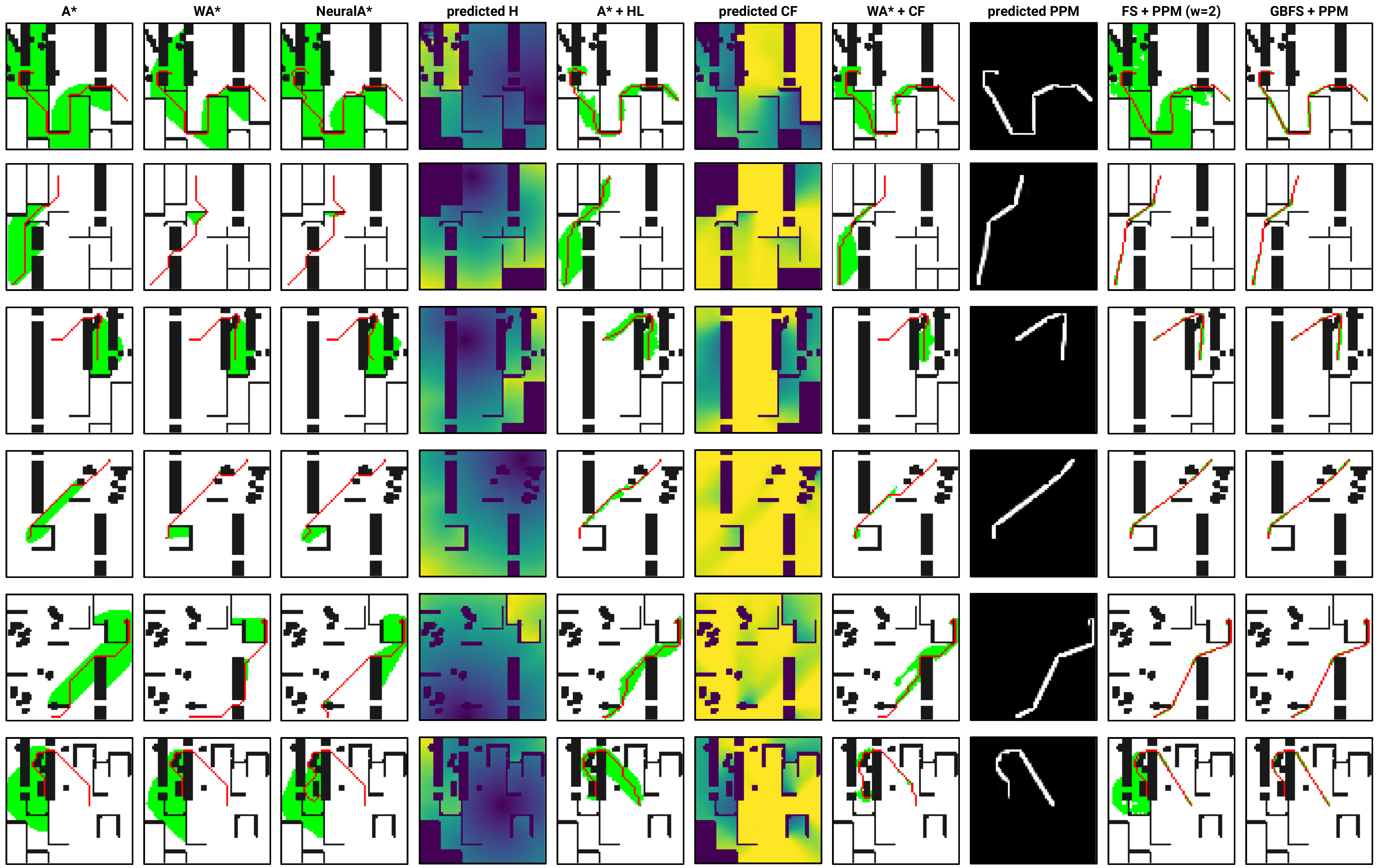}
    \caption{Randomly picked samples from the test part of our dataset.}
    \label{fig:solved_instance_examples}
\end{figure*}

As for the expansions ratio, the following trends can be observed. The performance of WA* and NeuralA* degrades when the hardness grows. Contrary, the performance of the other approaches does not change or even improves with growing hardness. The notable exception is FS+PPM which performance is very good when the hardness is lower than 2 and is inferior when the hardness exceeds this mark. This can be explained by the nature of the Focal Search, which was provided with the sub-optimality threshold of 2 in our experiments. FS hits the sub-optimality bound on the hard instances and is forced to expand redundant nodes with the lower $f$-values to rise the value of $f_{min}$ in $OPEN$ (which is needed to continue to progress towards the goa)l. Indeed, if the sub-optimality factor was set to a higher value, the drop of the performance would not be thus pronounceable, which is confirmed by the results of GBFS+PPM.

Fig.~\ref{fig:det_res} shows the examples of the predictions and the search results for the random pathfinding instances we were experimenting with.

\section{Out-of-the-distribution Experiments}
\label{appendix:out_of_distribution}

\begin{figure}
    \centering
    \includegraphics[width=\columnwidth]{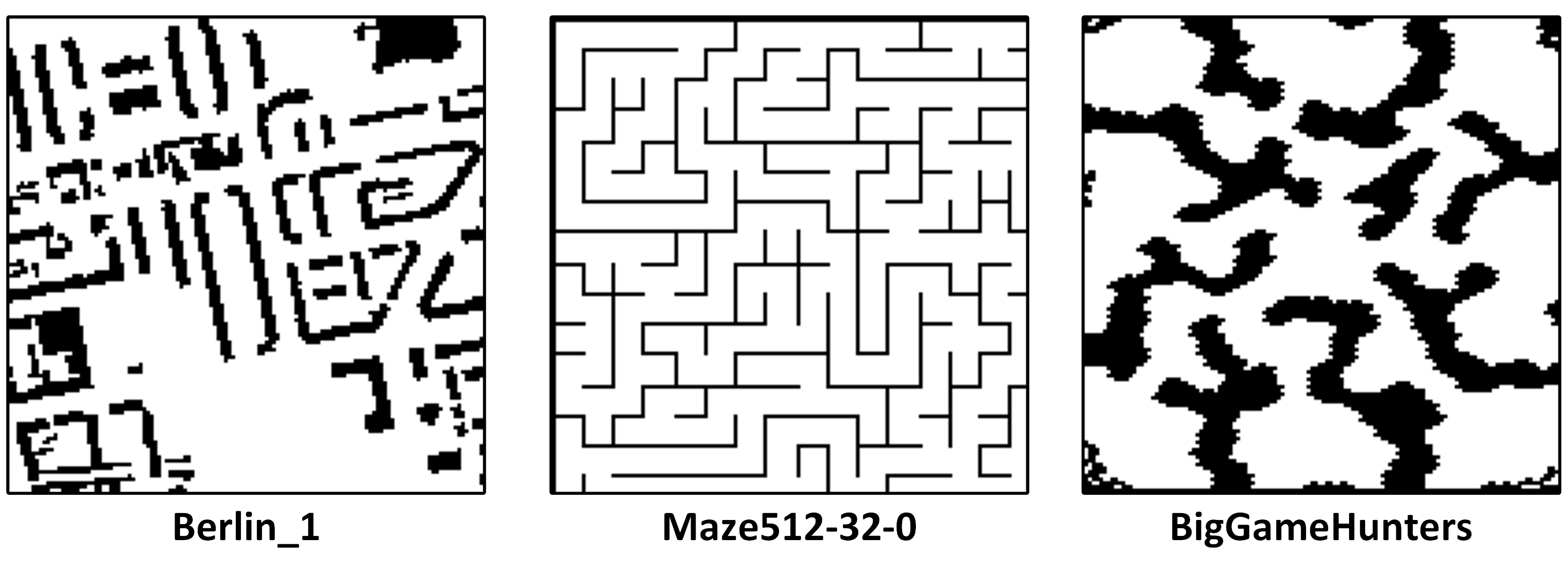}
    \caption{Maps comprising the out-of-distribution dataset.}
    \label{fig:out_maps}
\end{figure}

Besides the main dataset, we have also created an out-of-distribution dataset that consisted of three different maps taken from the MovingAI benchmark~\cite{sturtevant2012benchmarks}: \texttt{Berlin\_1} (City), \texttt{maze512-32-0} (Maze) and \texttt{BigGameHunters} (Game), see Fig.~\ref{fig:out_maps}. Each of these maps was scaled to two different sizes: $64\times64$ and $128\times128$. There were randomly generated $1,000$ instances per each map and size. As before, the instances with hardness less than $1.05$ were excluded from the experiments. 

We used this dataset to evaluate how the suggested in the paper learnable planners (and their competitors) perform when solving instances that are substantially different in topology and size from the ones used for learning. None of the maps from this dataset was used for training, i.e. these maps were produced tu the planners only at at the test phase.

Table~\ref{tab:out_res} presents the aggregated results. As expected the performance of all the learnable planners is worse compared to the main experiments. Still, the best results in terms of expansion ratio is achieved by one of our planners, i.e. GBFS+PPM (and the result of FS+PPM is very close).

\begin{figure}[t]
    \centering
    \includegraphics[width=\columnwidth]{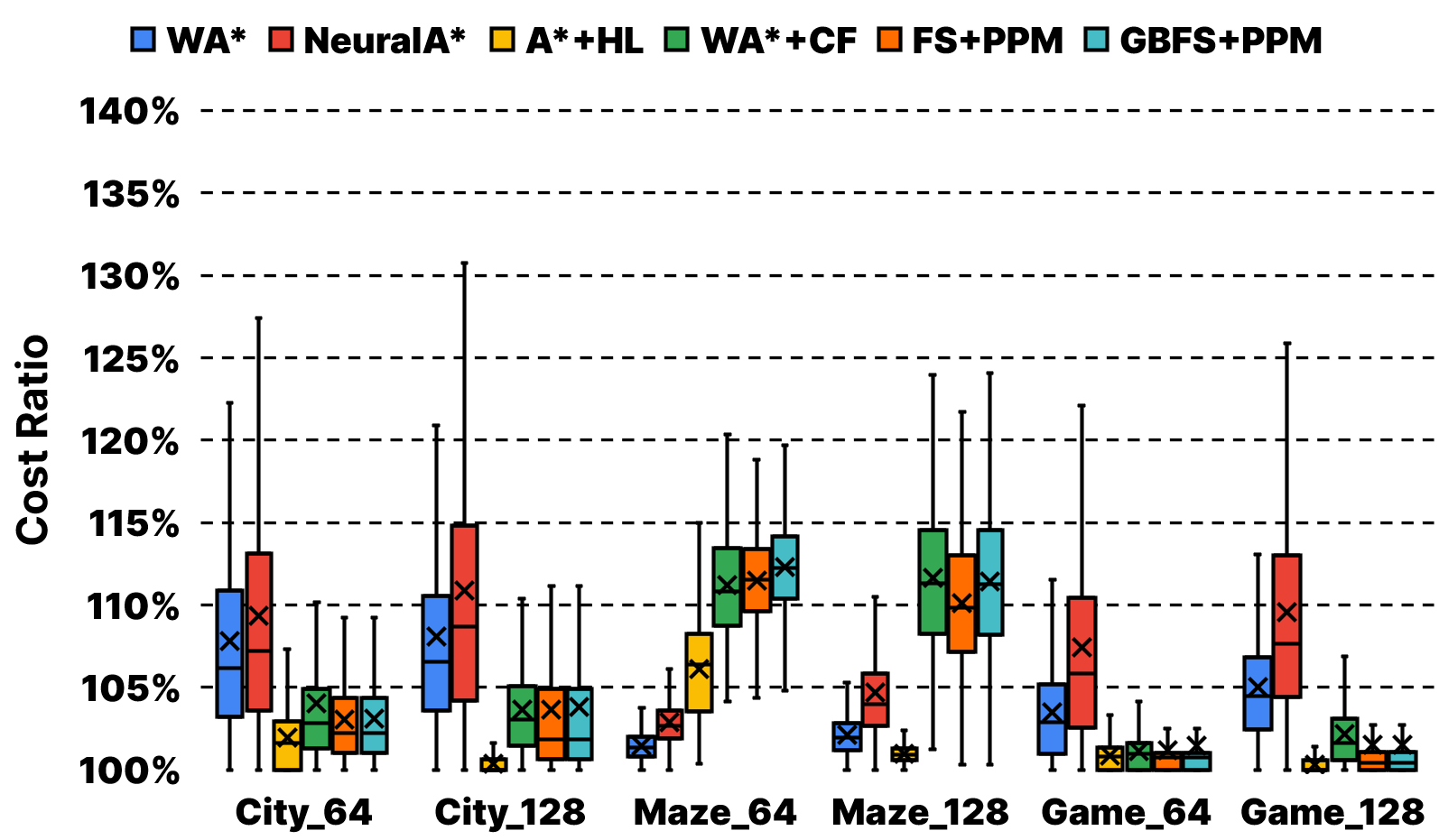}
    \caption{Box-and-whiskers plots of the cost ratio (out-of-the-distribution dataset).}
    \label{fig:out_det_res1}
\end{figure}

\begin{figure}[t]
    \centering
    \includegraphics[width=\columnwidth]{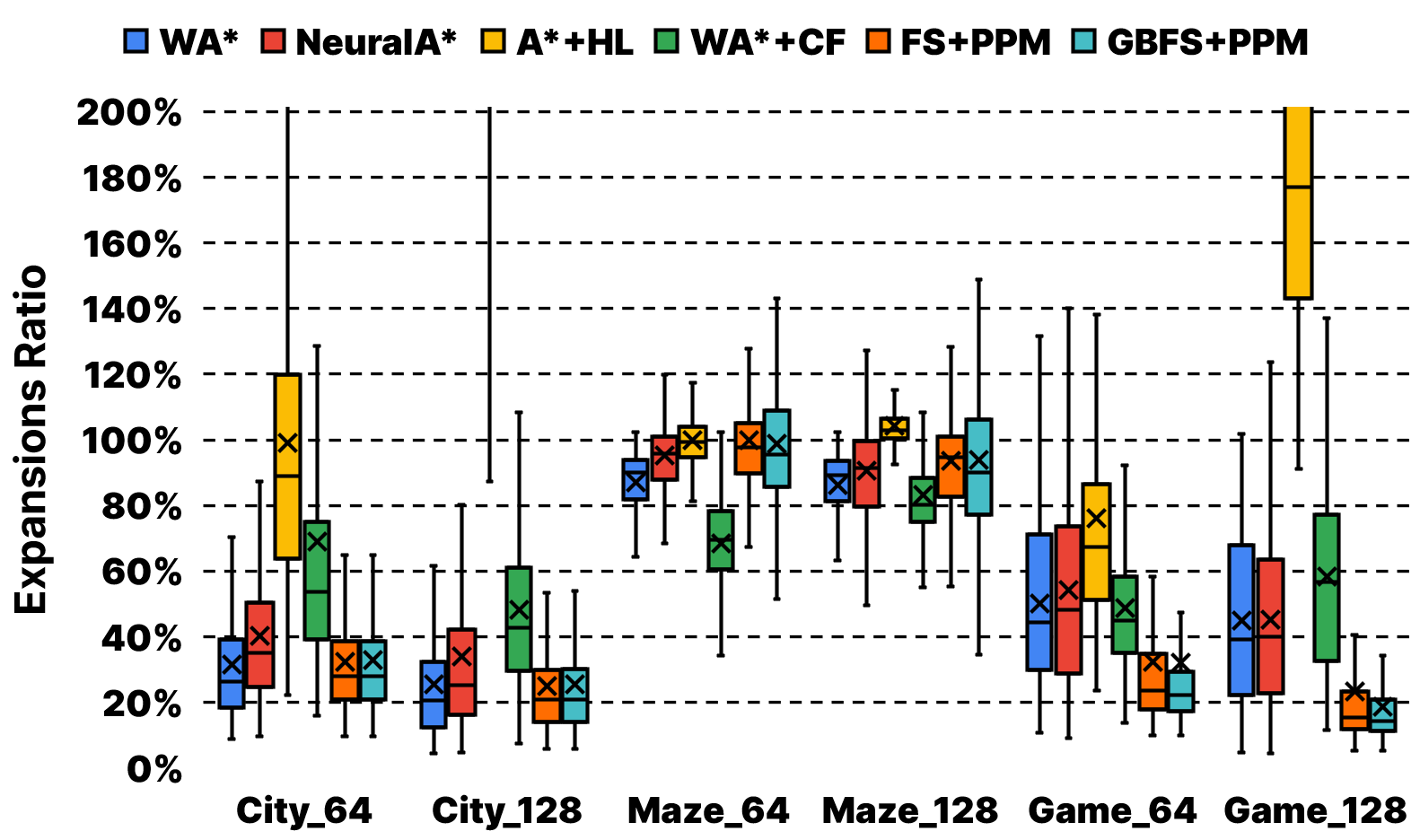}
    \caption{Box-and-whiskers plots of the expansions ratio (out-of-the-distribution dataset).\\ \\ \\ \\ \\ \\ \\ \\ \\ \\ \\}
    \label{fig:out_det_res2}
\end{figure}

The detailed box-and-whiskers plots for cost and expansions ratios for this experiment are depicted in Fig.~\ref{fig:out_det_res1},~\ref{fig:out_det_res2}. As one can note the results for \texttt{maze} map differ significantly, especially from the expansions ratio perspective. This can be explained by the fact that this type of maps is extremely hard to solve due to the arrangement of the blocked areas that do not form separate obstacles that can be circumnavigated. For the two other type of maps, however, the results are similar to the ones observed on our main dataset, i.e. FS+PPM and GBFS+PPM are able to reduce the amount of expansions up to a factor of 4 while producing just a slight overhead in terms of solution costs.

Overall, the observed results provide a firm ground to claim that the the suggested approaches have strong generalization capabilities and perform well on the out-of-the-distribution instances (at least until the topology of such instances is too complex, like in the case of mazes).

\begin{figure*}[t!]
    \centering
    \includegraphics[width=0.9\linewidth]{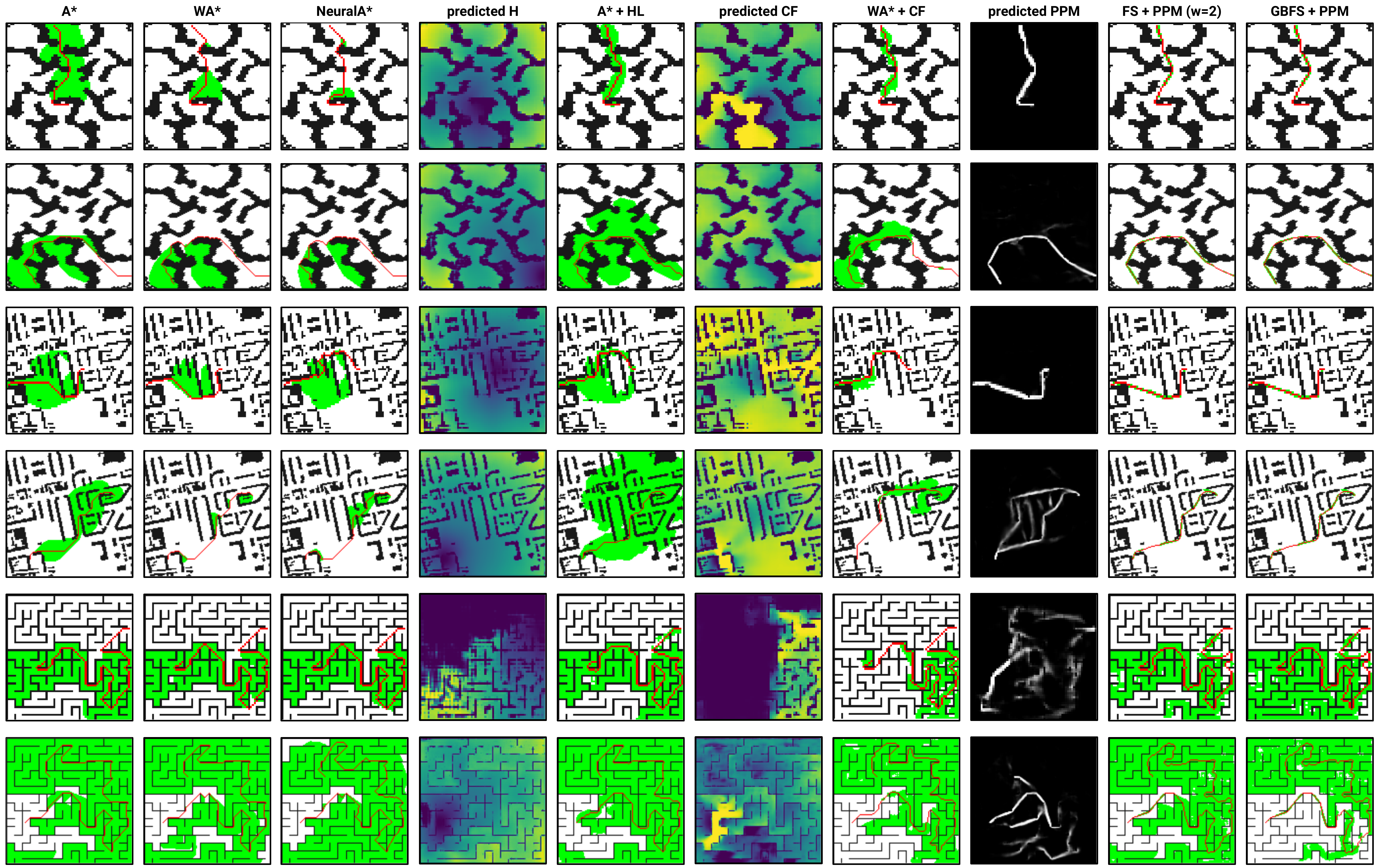}
    \caption{Randomly picked samples from the out-of-distribution dataset (no additional training was performed on this dataset).}
    \label{fig:ood_examples}
\end{figure*}

Figure~\ref{fig:ood_examples} shows the examples of the predictions and the search results for random instances from the main and the our-of-the-distribution datasets.

\newpage

\begin{table}[t]
    \centering
    \resizebox{.99\columnwidth}{!}{%
    \begin{tabular}{c|ccc}
    &Optimal Found &Cost &Expansions\\
    &Ratio ($\%$) $\uparrow$ & Ratio ($\%$) $\downarrow$ & Ratio ($\%$) $\downarrow$\\
    \hline
    A*& 100 & 100 & 100\\
    WA*& 8.13&	104.31	$\pm$4.76&	57.52	$\pm$30.72\\
    Neural A*& 3.24	&107.10	$\pm$6.77&	63.08	$\pm$34.63\\
    A*+HL& \textbf{29.02}&	\textbf{101.90}	$\pm$2.72&	148.94	$\pm$136.95\\
    WA*+CF&10.61&	106.10	$\pm$5.59&	63.64	$\pm$36.31\\
    FS+PPM&18.66&	105.62	$\pm$5.61&	55.06	$\pm$39.57\\
    GBFS+PPM&18.59&	106.12	$\pm$6.54&	\textbf{54.33	$\pm$47.24}\\

    \end{tabular}%
    }
    \caption{Experimental results on out-of-distribution dataset. Values before $\pm$ indicate the average, while after $\pm$ -- the standard deviation.}
    \label{tab:out_res}
\end{table}

\vfill
\end{appendices}
\end{document}